\documentclass[runningheads]{llncs}
\usepackage{graphicx}
\usepackage{tikz}
\usepackage{comment}
\usepackage{amsmath,amssymb} 
\usepackage{color}
\newcommand{\etal}{\textit{et al. }}

\usepackage{hyperref}

\begin{document}
\pagestyle{headings}
\mainmatter
\def\ECCVSubNumber{90}  

\title{WDRN : A Wavelet Decomposed RelightNet for Image Relighting} 

\titlerunning{WDRN}
\author{Densen Puthussery\thanks{Equal contribution} \and Hrishikesh P.S.$^\star$ \and Melvin Kuriakose \and Jiji C.V.}
\authorrunning{D. Puthussery et al.}
%
\institute{College of Engineering, Trivandrum, India\\
\email{$\{$puthusserydenson@, hrishikeshps@, memelvin@, jijicv@$\}$cet.ac.in}}
\maketitle

\begin{abstract}
The task of recalibrating the illumination settings in an image to a target configuration is known as relighting. Relighting techniques have potential applications in digital photography, gaming industry and in augmented reality. In this paper, we address the one-to-one relighting problem where an image at a target illumination settings is predicted given an input image with specific illumination conditions. To this end, we propose a wavelet decomposed RelightNet called WDRN which is a novel encoder-decoder network employing wavelet based decomposition followed by convolution layers under a muti-resolution framework. We also propose a novel loss function called gray loss that ensures efficient learning of gradient in illumination along different directions of the ground truth image giving rise to visually superior relit images. The proposed solution won the first position in the relighting challenge event in advances in image manipulation (AIM) 2020 workshop which proves its effectiveness measured in terms of a Mean Perceptual Score which in turn is measured using SSIM and a Learned Perceptual Image Patch Similarity score.

\keywords{Gray Loss, Illumination, Relighting, Wavelet}
\end{abstract}

\section{Introduction}

\par
\begin{figure}
\centering
\newcommand\x{0.24}
\newcommand\scale{0.15}
  \begin{minipage}{\x\linewidth}
		\begin{center}
		\includegraphics[scale=\scale]{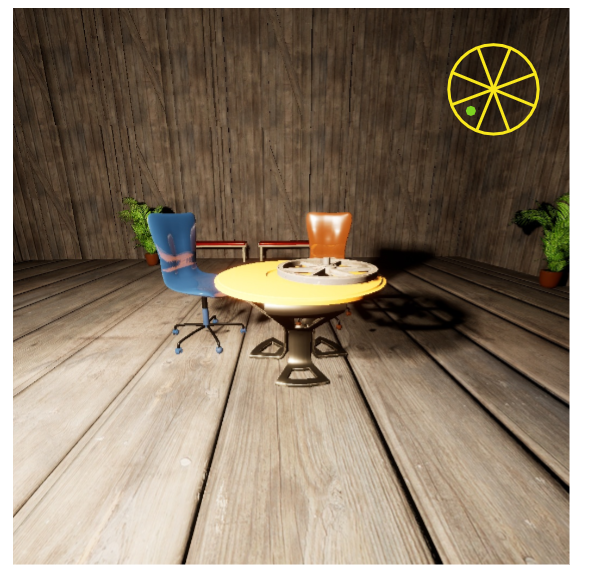}
		\fontsize{8}{12pt}\selectfont (a) South-West
		\end{center}
  \end{minipage}
  \begin{minipage}{\x\linewidth}
		\begin{center}
		\includegraphics[scale=\scale]{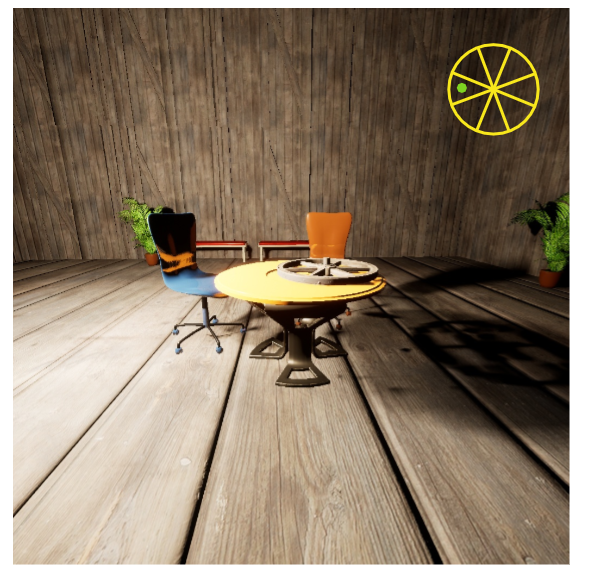}
		\fontsize{8}{12pt}\selectfont (b) West
		\end{center}
  \end{minipage}
  \begin{minipage}{\x\linewidth}
	\begin{center}
	\includegraphics[scale=\scale]{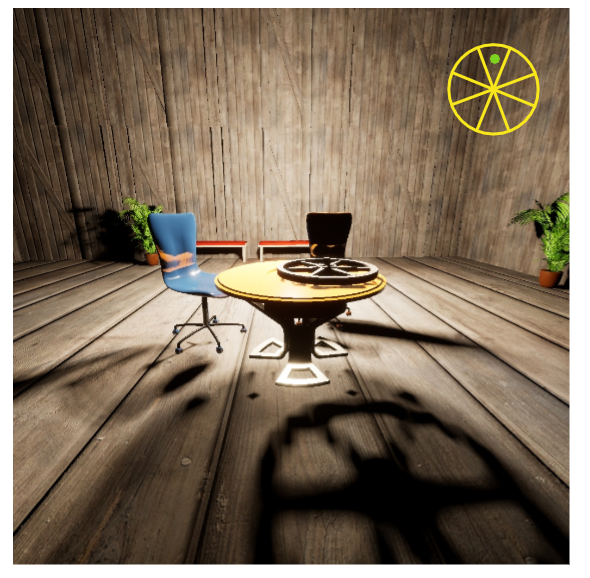}
	\fontsize{8}{12pt}\selectfont (c) North
  \end{center}
  \end{minipage}
  \begin{minipage}{\x\linewidth}
		\begin{center}
		\includegraphics[scale=\scale]{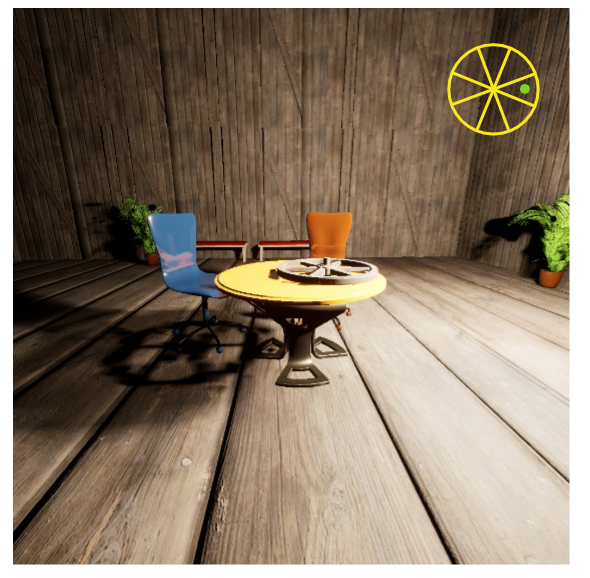}
		\fontsize{8}{12pt}\selectfont (d) East
		\end{center}
  \end{minipage}
\caption{An example of scene relighting for a change in illuminant position. It can be observed that the shadows caste in a, b, c and d are quite different from each other owing to the different relative position of objects and illuminant in each case. Also, the gradient in brightness is different in each case since the region in the scene proximal to the light source is different.}
\label{fig:intro_pic}
\end{figure}

 The task of recalibrating the illumination settings of an acquired image is widely known as image relighting. Relighting is an emerging technology owing to its applications in augmented reality (AR) and also in casual digital photography. Relighting enabled AR can bring about  great changes in the way one perceives digital experiences like online shopping, online teaching, etc. For example, one may wish to visualize whether furniture to be purchased online is suitable for the room. Since the ambient lighting conditions like the direction of illumination, brightness, color temperature etc. may vary from user to user, adaptability to the same is required in the AR visualization tool. Such adaptability can be realized by integrating relighting techniques in the online platform using the AR tool. In first and third person gaming, the ambient lighting of a scene is highly dynamic and changes with time of the day, viewpoint of the avatar etc. There is a scope for using relighting techniques to quickly render the scene graphics to drive the gameplay with higher number of frames per second. Fig. \ref{fig:intro_pic} shows a simple case of relighting where the appearance of the scene is changed drastically when the illuminant is positioned at different azimuthal angles.

\par In digital photography, relighting techniques are used to enhance the perceptual quality of an image. In natural images of outdoor scenes, it is often difficult to control the illumination. Diverse factors affect natural illuminance like time of the day, weather, clouds, objects in the vicinity etc. Due to these and other factors, it is common that outdoor images are poorly lit. Many modern cameras offer the flexibility to control the image lighting by adjusting the shutter speed, aperture, ISO sensitivity etc. However, such tweaks usually require professional expertise and are prone to degradation like blur, grains etc. 

\par The outdoor images usually have  uniform illumination in daytime as sunlight is far-field and heavily scattered. However, indoor images usually have a non-uniform illumination as the objects close to the light source are considerably lit in contrast to the ones far away. The location, directionality and properties of illuminant dictate the appearance of natural indoor images. The formation of shadows and general gradient of illumination are controlled by the location of illuminants in the room. 
\par Additionally, the nature of object shadow like its size, position etc. varies with the location of the light source. Similarly, the illumination pattern produced by a directional source is quite different from that of an omni-directional light source. Besides, the properties of illuminant like its color temperature, spectral power distribution etc. affect the visual quality of an indoor image. The above factors hold true even for outdoor images when natural lighting is not present. Although indoor photography is flexible, as we have a control on these factors, it is only feasible on a professional scale, like in a digital studio. In a home environment and in casual photography, the location and type of illuminant are mostly fixed and there is little control on these aspects. Relighting finds its applications in areas like these where one would like to change the illumination setting of an image without putting much physical effort or using specialized tools.

\par Another area of image manipulation which necessitates relighting is digital image montaging. In image montaging, a certain portion of an image is replaced with a crop taken from a different image. Multiple images can also be fused in a similar manner to generate a montage. For seamless and visually appealing results in a montage, the illumination in images being combined must be same. Since the images used for montaging are usually unrelated, their illumination setting could be different. In this scenario, relighting techniques can be employed to translate the images into the final illumination setting and then apply montaging for superior results.

 In this work, we address a special case of the relighting problem where we describe the solution we proposed as part of the challenge event on one-to-one relighting in Advances in Image Manipulation(AIM) 2020 workshop \cite{elhelou2020aim}. The task of the challenge was to develop a solution that recalibrates the illumination setting of an input scene to a given target setting. To this end, a deep convolutional neural network (CNN) that efficiently learns the illumination setting of the target domain is proposed.
 The contributions of the proposed work are :
\renewcommand{\labelitemi}{\textbullet}
\begin{itemize}
\item A wavelet decomposed encoder-decoder network to solve the relighting problem that can effectively translate an image from a source illumination setting to a target setting.
\item A novel training loss term called gray loss that drives the network to learn the illumination gradient in target domain images.
\item  Introduced pixel shuffler operations in wavelet based encoder-decoder network for fast training and inference.
\end{itemize}
Rest of the paper is organized as follows : In Section \ref{sec:related_wok} we review related works and in Section \ref{section:proposed_method} we describe the proposed methodology. Section \ref{section:experiments} details our experiments, Section \ref{sec:result_analysis} presents our result analysis and in Section \ref{section:ablation_studies} we describe our ablation studies. Finally, Section \ref{section:conclusions} concludes the work.

\section{Related Work}
\label{sec:related_wok}
Here we first review the image enhancement techniques, both using conventional and deep learning approaches where the enhanced image is obtained through some form of illumination adjustment. Further, we discuss some of the recently proposed relighting techniques using deep networks.
\subsection{Image Enhancement}
\subsubsection{Conventional Methods}
\par
Smartphones and casual photography using these
devices have brought an increased demand for methods based on various image
manipulation techniques like photo enhancement. Image enhancement is one of
the fundamental problems in the field of computer vision starting with methods
like histogram equalisation for contrast enhancement. Retinex theory of color
vision \cite{retinex} by Edwin H Land inspired many methods like \cite{retinex2}\cite{retinex_1} that considers images as the pixel-wise product of reflectance and illumination. These works treat image enhancement problem as an illumination estimation problem, where the illumination component is used to enhance the input images. These works were
only able to generate very inferior results because of the high non-linearity across
the channels and the spatial sensitivity of colour in the image.
\par
\subsubsection{Deep learning based Methods}
\par
Most of the recent works on photo enhancement is learning based and the first dataset used for the purpose was MIT-Adobe FiveK \cite{MIT_dataset} introduced by Bychkovsky \etal The dataset contains five sets of $5000$ input-output pairs. Each set is a retouched (using Adobe Lightroom) version of the same input image by different professionals.
The work was used to address general tone adjustment rather than enhancing an underexposed image.
\par
Lore \etal \cite{LLNet} proposed an auto-encoder architecture for denoising and brightening the low-light images.
Many Generative Adversarial Network (GAN) based networks were also developed for image enhancement. Chen \etal \cite{Deep_photo_enhance} proposed a method that uses a two-way GAN architecture. The network transforms the input image to an enhanced image with characteristics of a reference image.  Ignatov \etal \cite{WEPSE}, proposed weakly supervised (no exact image pair) GAN based network that enhances images that are taken using mobile phones to DSLR quality images. The network used DPED dataset \cite{DPED} along with many other unpaired HD images.
\par
Wang proposed a learning based method \cite{wang} that enhanced under exposed images using end-to-end CNN based model. The network used an encoder-decoder architecture, where the encoder was used to extract the local features like contrast, detail sharpness, shadow, highlight etc and global features such as color distribution, average brightness and scene category. For driving the network to learn illumination mapping from under-exposed to the enhanced images, they use three loss functions, smoothness loss on the illumination and color and
reconstruction loss on the enhanced image. The network was trained on
a novel dataset with 3000 under-exposed images and its ground truth.
\subsection{Image Relighting}
\label{sec:Image Relighting}
\par
Most of the above mentioned methods cannot remove or change the illumination
setting of an input image; it can only modify the effects of the existing illumination. When it comes to image relighting rather than the overall enhancement,
the work focuses on predicting a target illumination setting (light direction and
colour temperature) from an input with a different illumination setting.
\par
One-to-one relighting can be considered as a special case of image relighting,
where the task is to manipulate an input image that was captured under certain
illumination settings (light source position, direction and color temperature) to
make it look like it was taken under different settings.
\par
Debevec \etal proposed a technique \cite{reflectance_human_face} for rendering the human face images from varying
viewpoints and direction of illumination, similar to the problem addressed in
our work. Here, they collected images of human face from different viewpoints
under diverse direction of illumination. A reflectance function of the skin was
modeled to estimate the image when the target viewpoint is different from the
input. Their network was able to give considerable performance but it required
hundreds of images with the stationary subject under a controlled illumination setting. Hence,
they were unable to provide a solution for single RGB image for an unknown
object in an unconstrained environment as in one-to-one relighting problem.
\par
Xu \etal proposed a CNN based method \cite{relightnet_a} to relight a scene under a new illumination based on five images captured under pre-defined illumination setting. Unlike exploiting similarity in a single light transport function as in \cite{reflectance_human_face} they try to estimate a non-linear function that generalises the estimation of the above mentioned problem using deep learning based training. Along with three channels(RGB)
of the five fused images, they also add two extra channels along with the image, which are
2D coordinates of the light source direction. This method still requires five sparse
samples of the same scene in order to predict the scene from a novel light setting.

\par
Indirectly addressing this problem, Sun \etal proposed a CNN based approach \cite{Single_image_potrait_relighting}
to relight portrait images that were taken on mobile cameras into user defined
illumination setting. They also used a encoder-decoder architecture where the
input illumination is predicted and the required illumination of the target is injected
at the bottleneck layer between the encoder and the decoder. The work was able
to develop a function that can predict diverse illumination but their work was
limited to portrait images of human faces.
\par
The proposed method uses wavelet based end-to-end CNN architecture inspired
from Multi-level Wavelet-CNN (MWCNN) \cite{mwcnn} by Liu \etal to learn a mapping function
that relights a scene without modeling for the geometry or the reflectance. MWCNN is a fully convolutional encoder-decoder network that was proposed as a general methodology for image restoration. The winners of NTIRE 2020 Challenge on image demoireing \cite{NTIRE_2020_CVPR_Workshops} used a method inspired from MWCNN which shows its competence.
In this work, to relight an input image to a given target illumination settings, we propose a deep convolutional network using wavelet decomposition followed by convolution layers at various scales utilizing novel loss functions.

\nocite{vidit_based, light_transport, reflectance_modeling}

\section{Proposed Method}
\label{section:proposed_method}
\subsection{Problem Formulation}
\par Under the assumption of a distant illumination source, the scene to be relit under a target direction can be formulated from the light transport function $\lambda(i,\theta)$ and the incident source illuminations from direction $\theta$ as :
\begin{equation}
    Y_{i} = \int\lambda(i,\theta)I(\theta)d\theta
\end{equation}
where $I(\theta)$ is the radiance of the incident illumination from direction $\theta$ and $i$ is the target image pixel. In a fundamental relighting problem, given multiple images of the scene acquired under varying $\theta $, $\lambda(i,\theta)$ can be estimated and then image corresponding to a new value of  $\theta$ can be rendered \cite{relightnet_a}.
\par 
The problem that we discuss in this paper is slightly different, where we describe the solution proposed as part of the one-to-one relighting challenge at Advances in Image Manipulation (AIM) 2020 workshop \cite{elhelou2020aim}. The task of the challenge was to develop a solution that recalibrates the illumination setting of an input scene to a target setting. The illumination setting in the challenge refers to two aspects - position and color temperature of the light source. Thus for the given problem, the input image is characterized by a fixed light source direction $\theta_1$ and a fixed color temperature $T_1$ while the target image is characterized by a different direction $\theta_2$ and color temperature $T_2$. We employ a deep convolutional neural network (CNN) to learn the complex function $F(\theta_1, \theta_2,T_1,T_2)$ which can render the given scene into the new settings $\theta_2, T_2$.

\subsection{Proposed Wavelet Decomposed RelightNet (WDRN)}
\subsubsection{Overview}
\par The proposed method uses a multi-level encoder-decoder based network that processes the image at different spatial resolutions. The encoder section is used to extract the local features like contrast, sharpness, shadow and global features such as color distribution, brightness, and semantic information. The encoder learns the illumination mapping of the input based on the extracted features. The decoder reconstructs the relit images from the encoder output by progressively upsampling the feature maps to the resolution of the target image.

\par
Also, feature information from the encoder level is forwarded into the decoder level that operates at the same spatial resolution. The information in the decoder is the image context and the forwarded information is the local and global features. By fusing local  and contextual information, the target illumination setting is injected into the input image within the decoder. The detailed description of the encoder and decoder sub-net is given in the following sub-sections.

\par
\begin{figure}
    \centering
    \includegraphics[scale =0.2]{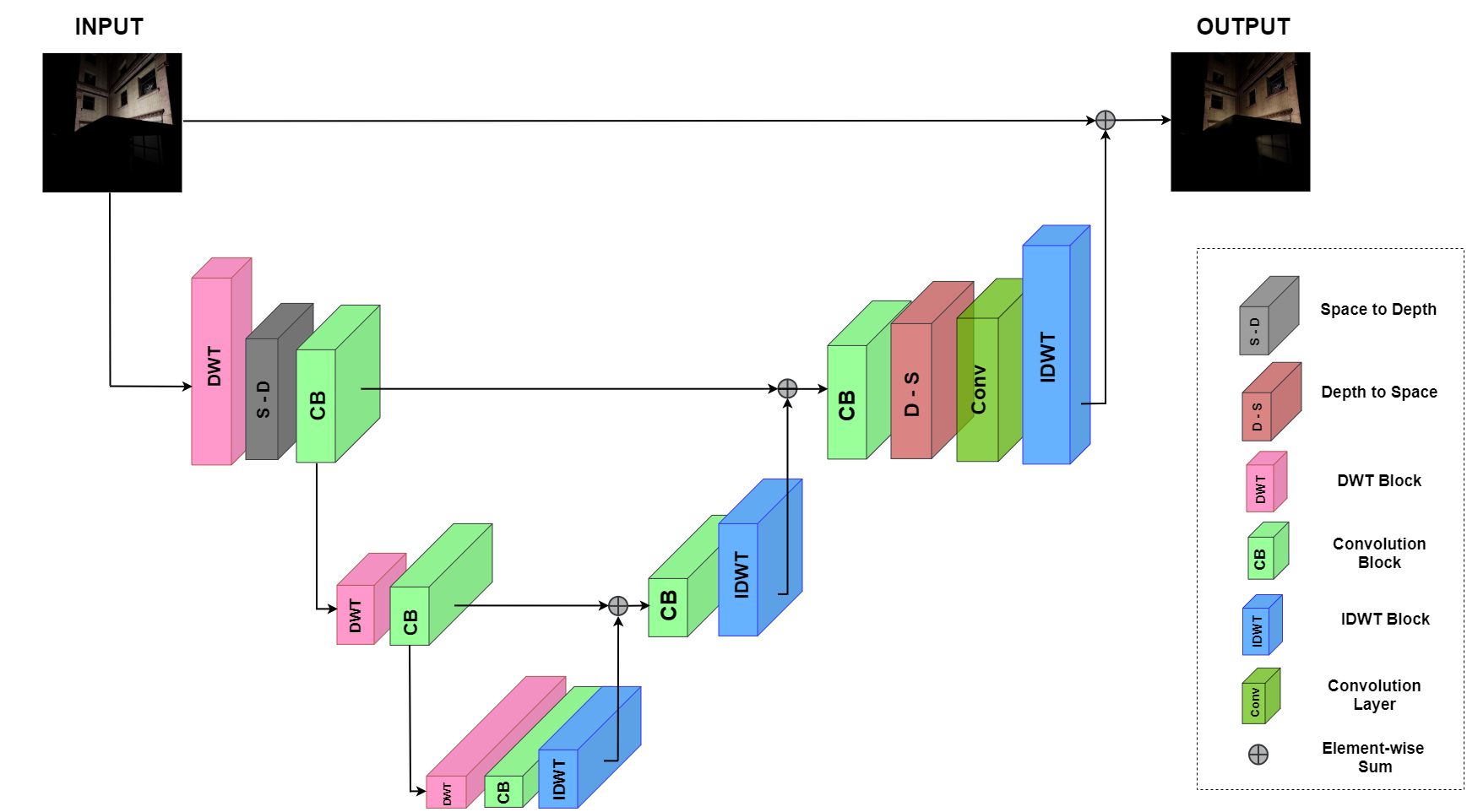}
    \caption{The proposed WDRN architecture. There are 3 different processing levels in the network. The subsampling and interpolation are done using DWT and IDWT respectively. The operation of a simple DWT-IDWT based encoder-decoder network is depicted in Fig. \ref{fig:dwt}. Convolution blocks are used for feature extraction at each level and the block is expanded in Fig. \ref{fig:conv_block}
    }
    \label{fig:wdrn}
    \end{figure}

 The network is termed Wavelet Decomposed RelightNet (WDRN) because it employs wavelet decomposition to process the image at different scales within the encoder-decoder architecture. WDRN is inspired by the work of multi-level wavelet CNN (MWCNN) for image restoration proposed by  Liu \etal \cite{mwcnn}. The ability of wavelet transform to obtain a large receptive field without information loss or gridding effect was shown in their work. Fig. \ref{fig:wdrn} depicts the proposed WDRN architecture for relighting.
\begin{figure}
    \centering
    \includegraphics[scale=0.25]{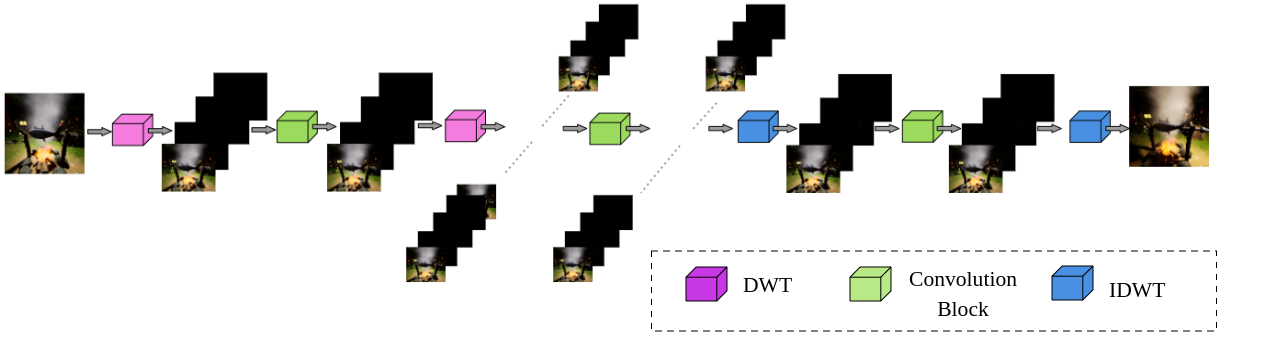}
    \caption{The operation of a simple DWT-IDWT based encoder-decoder network. Here the input is subsampled and divided into 4 sub-bands and these sub-bands are further processed using convolution block. The same step is repeated in the subsequent levels of the encoder. In the decoder, based on these feature maps information the relit image is reconstructed into required spatial resolution.}
    \label{fig:dwt}
\end{figure}
\begin{figure}
    \centering
    \includegraphics[scale=0.18]{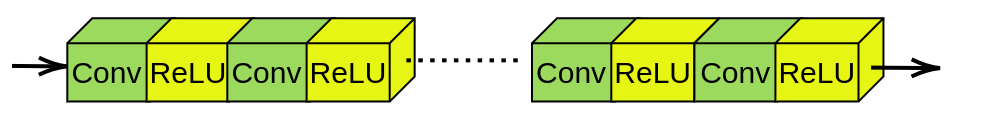}
    \caption{Convolution block contains a series of convolutions and ReLU activations that are stacked in sequence. The number of these conv-ReLU blocks varies in different encoder and decoder levels.}
    \label{fig:conv_block}
\end{figure}

\subsubsection{Encoder sub-net}
\par The encoder sub-net operates at three different image scales or spatial resolutions and hence is a multi-level network. In each level, the input feature map is decomposed into four sub-bands using discrete wavelet transform (DWT) based subsampling followed by a convolution block for feature extraction. Here, 2D Haar wavelet has been used for decomposing the input to its sub-bands since it is the simplest of its kind. Other types of wavelets can also be employed here on a trial and error basis as it is difficult to theoretically prove the most suitable wavelet for this operation.  The main advantage of this processing step is that a high receptive field is obtained in the network, similar to that with a dilated convolution. There are no trainable parameters in this decomposition step unlike in a subsampling convolution. The network will also benefit from frequency and spatial localization capabilities of the wavelet transformation. Fig. \ref{fig:dwt} depicts the operation of a simple DWT-IDWT based encoder-decoder network. Additionally, in the first encoder level, a space-to-depth transform with pixel shuffler is applied on the decomposed sub-bands to generate a feature map of quarter the area of input and four times the number of channels. There are no trainable parameters in this downscaling operation, rather, it is a simple rearrangement of the subpixels. One can now perform subsequent processing of the original features at a smaller resolution which makes the overall network computationally  efficient.

\par The convolution block in each encoder level is a series of convolutional layers followed by ReLU activation as depicted in Fig. \ref{fig:conv_block}. The convolution block is used for feature extraction from the input to the block. It learns the local features of the image like contrast, sharpness, shadow etc. and global features such as color distribution, brightness, and semantic information. The number of convolutional layers and filters are different for each level of the encoder. In level one, there are four convolution layer with $16$ filters in each layer. Similarly, in level two there are four convolution layers with $6$ filters. In level three there are seven convolution layers each with $256$ filters. In the contraction path of the encoder, the filter size is progressively increased to obtain a rich representation of lower scale features.
\subsubsection{Decoder sub-net}
\par Similar to the encoder, the expansion path sub-net or decoder is also multi-level and has three different scales of operation. Each level in the decoder is constituted by an inverse discrete wavelet transform (IDWT) based interpolation followed by a convolution block for feature aggregation. The feature output of third level of encoder is the input to the first level in the decoder. In each level, the input features are assumed to be four sub-bands of a wavelet decomposition. With this assumption, the IDWT is computed on the input feature set to interpolate the features to twice their spatial resolution and a quarter of the total input channels. The features in the expansion path represent the contextual information of the image. Since the local and global features are present in different encoder level outputs, these features can be carried forward to the decoder. This is achieved by directly adding the interpolated decoder features with the encoder level outputs with the same spatial resolution in an element-wise manner. This output feature set is then processed through a convolutional block to gradually inject the target domain illumination setting. 
\par Similar to encoder, the convolution block in a decoder level is constituted by convolutional layers followed by ReLU activations. Convolution block in Level one of the decoder has four convolution layers of $64$ filters each. In level two, there are four convolution layers of $16$ filters each. At the output of the second level, a depth-to-space transform with pixel shuffler is employed which serves as the inverse operation of space-to-depth at the input. The third level is constituted only of a single IDWT operation. The last IDWT interpolation generates a three channel feature map which is added to the input image to generate the relit image of target illumination settings. As the last IDWT operation should produce a $3$-channel image, the total number of channels at the input of level three IDWT should be $12$. Since the output of depth-to-space transform has $16$ channels, a convolution layer with $12$ filters is placed after it to adjust the depth in subsequent layers. 
\par In general, for efficient illumination recalibration, the network should be able to establish the relationship between distant pixels. This can be realised by using highly dilated convolutions. But for large dilation factors, two adjacent pixels in the predicted feature map are calculated from the completely non-overlapping input feature set and hence leads to spatial information leakage and poor localization in the encoder levels. The proposed WDRN can achieve a high receptive field without this information loss. Moreover, in contrast with MWCNN, the training losses used in WDRN is tailored for the relighting problem. The details of the novel gray loss that we propose for perceptually superior results in relight problem and other losses that we used are detailed in the next section.

\subsection{Loss Functions}
\par
Network is trained based on three empirically weighted loss functions as shown in equation \ref{eq:total loss}.

\begin{equation}
    L_{total} =  \alpha L_{MAE}+\beta L_{SSIM}+\gamma L_{gray}
    \label{eq:total loss}
\end{equation}
The MAE loss is the mean absolute error or the $L_1$ distance between the ground-truth and the predicted images. It is incorporated to generate high fidelity relit images. Mean squared error (MSE) loss was avoided because of the smoothening effect it introduced in our generated images. The MAE is given by :
\begin{equation}
    L_{MAE} = \frac{1}{W\times H\times C} \sum_{i=0}^{W-1}\sum_{j=0}^{H-1}\sum_{k=0}^{C-1}\left | Y_{i,j,k} - \hat{Y}_{i,j,k}\right |
    \label{eq:MAE loss}
\end{equation}
where, $W$, $H$ and $C$ are the width, height and number of channels of the output, $Y$ is the ground truth image and $\hat{Y}$ is the predicted image.
The structural similarity (SSIM) \cite{wang2004image}  between two images is a measure of their perceptual difference as SSIM incorporates contrast and luminance masking. A high dynamic range can reveal more details in both poorly and heavily lit regions in an image. Optimising for SSIM loss helps the network to render visually appealing images with better dynamic range. SSIM loss is formulated as :

\begin{equation}
    L_{SSIM} = 1- SSIM(Y,\hat{Y})
    \label{eq:SSIM loss}
\end{equation}

\subsubsection{Gray Loss}
\par In relighting problems where the objective is to change the general direction of lighting, the network should be able to recalibrate the gradients in illumination within the image. The objects closer to the target illuminant position should be heavily lit while the ones far away should be poorly lit. The MAE loss and SSIM loss can optimize for the general texture of the image, but not the gradients in illumination. Hence the enhancements like shadow recasting are poorly learned. A novel loss term called gray loss is hence proposed that can overcome these limitations. The proposed gray loss is the $L_1$ distance between blurred versions of the grayscale components of the relit and ground truth images. The texture details are smoothened out when the images are blurred, leaving behind the illumination information. Since much details are not present in the blurred image, this information is constituted by the general direction of illumination gradients. Thus gray loss ensures that the gradient in illumination along different directions of the ground truth image is learned by the network and generate visually superior results. Gray loss is formulated as :

\begin{equation}
    L_{gray} = \frac{1}{W \times H} \sum_{i=0}^{W-1}\sum_{j=0}^{H-1}\left | (\psi(Y))_{i,j} - (\psi(\hat{Y}))_{i,j}\right |
    \label{eq:gray_oss}
\end{equation}
where $\psi(.)$ is the Gaussian blur function used to smoothen the images.

\section{Experiments}
\label{section:experiments}
\subsection{Dataset}
The dataset used in the experiments is the Virtual Image Dataset for Illumination Transfer (VIDIT) \cite{vidit}. The dataset contains $390$ different scenes which is captured at $40$ different illumination settings ($8$ azimuthal angles and five different colour temperatures $2500$K, $4500$K etc.) with a total of $15,600$ images. We participated in track $1$ - one-to-one relighting in AIM 2020 challenge for Scene Relighting and Illumination Estimation. For the experiments as part of the challenge, we used $390$ image pairs from the dataset, where the input image has a fixed illumination setting $\theta_1$=North, $T_1$ = $6500$K and the target is set at a different illumination setting $\theta_2$ = East, $T_2$ = $4500$K. All the training images are of fixed size $1024\times1024\times3$. 
Out of the $390$ image pairs, $300$ image pairs were used for training, $45$ for validation and $45$ for testing. 
\subsection{Training}
\par

\begin{figure}
\centering
\newcommand\x{0.49}
\newcommand\scale{0.34}
  \begin{minipage}{\x\linewidth}
		\begin{center}
		\includegraphics[scale=\scale]{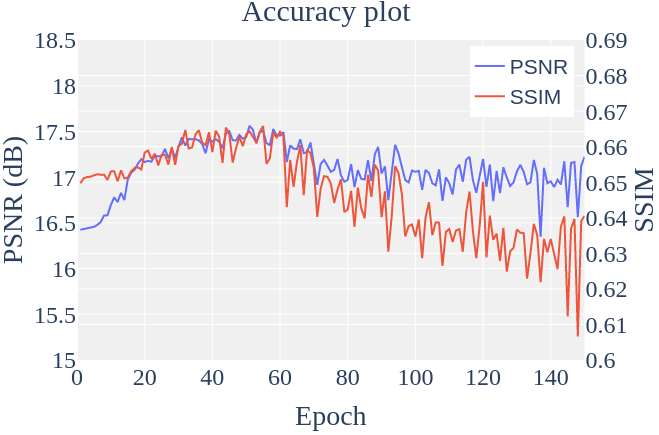}
		\fontsize{8}{12pt}\selectfont (a)
		\end{center}
  \end{minipage}
  \begin{minipage}{\x\linewidth}
		\begin{center}
		\includegraphics[scale=\scale]{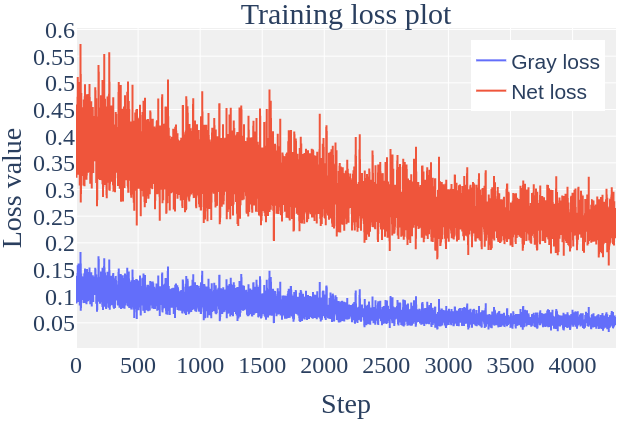}
		\fontsize{8}{12pt}\selectfont (b)
		\end{center}
  \end{minipage}
\caption{ Plots for (a) Accuracy  (b) Training Losses}
\label{fig:plots}
\end{figure}

The network was trained on mini-batches of size $10$. The model was trained for $150$ epochs and employed Adam optimiser with $\beta_1=0.9$ and $\beta_2=0.99$. The initial learning rate was $1e^{-4}$ which was then decayed by a factor of $0.5$ after every 100 epochs. The training was done on $1\times$ Tesla P100 GPU card with $16$ GiB memory. The proposed network has $6.4$ million trainable parameters. Training process took $2$ hours and testing time per image was $0.03$ seconds. The various training accuracy and loss plots are shown in Fig. \ref{fig:plots}. It can be inferred from the accuracy plot that the network overfits at around $60$ epochs.


\subsection{Evaluation Metrics}
In addition to the standard evaluation metrics like peak signal to noise ratio (PSNR) and SSIM, the performance of the proposed WDRN is evaluated using rather new perceptual metrics like Learned Perceptual Image Patch Similarity (LPIPS) \cite{zhang2018unreasonable} and mean perceptual score (MPS). Mean Perceptual Score (MPS) \cite{elhelou2020aim} is the average of the normalized SSIM \cite{wang2004image} and LPIPS score as shown in Eq.\ref{eq:MPS}
\begin{equation}\label{eq:MPS}
    MPS = 0.5 (S + (1-L))
\end{equation}
where $S$ is the average SSIM score on the test set, and $L$ is the average LPIPS score on the test set.

\section{Result Analysis}
\label{sec:result_analysis}
\begin{figure}
\centering
\newcommand\x{0.23}
\newcommand\scale{0.078}
  \begin{minipage}{\x\linewidth}
		\begin{center}
		\includegraphics[scale=\scale]{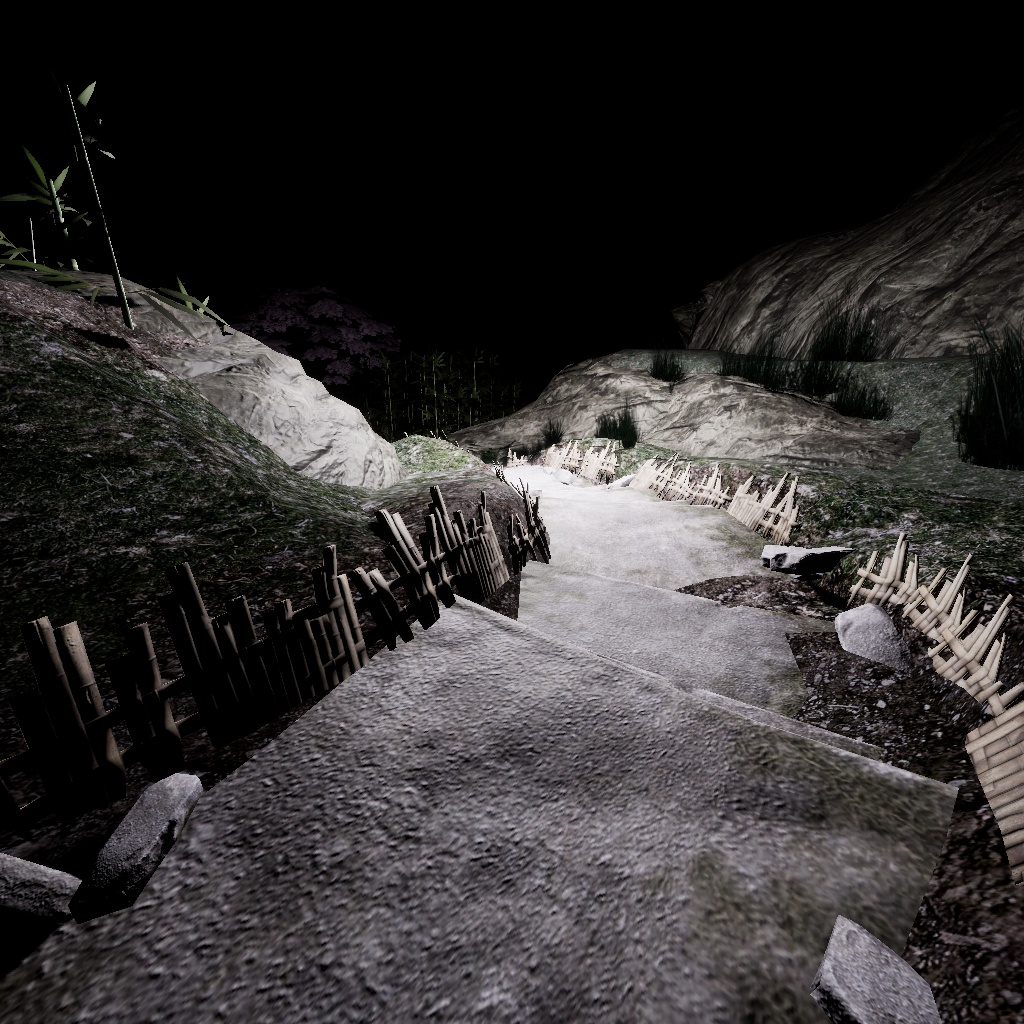}
		\fontsize{8}{12pt}\selectfont PSNR-$13.33$ dB \\SSIM-$0.63$
		\vskip 2pt
        \includegraphics[scale=\scale]{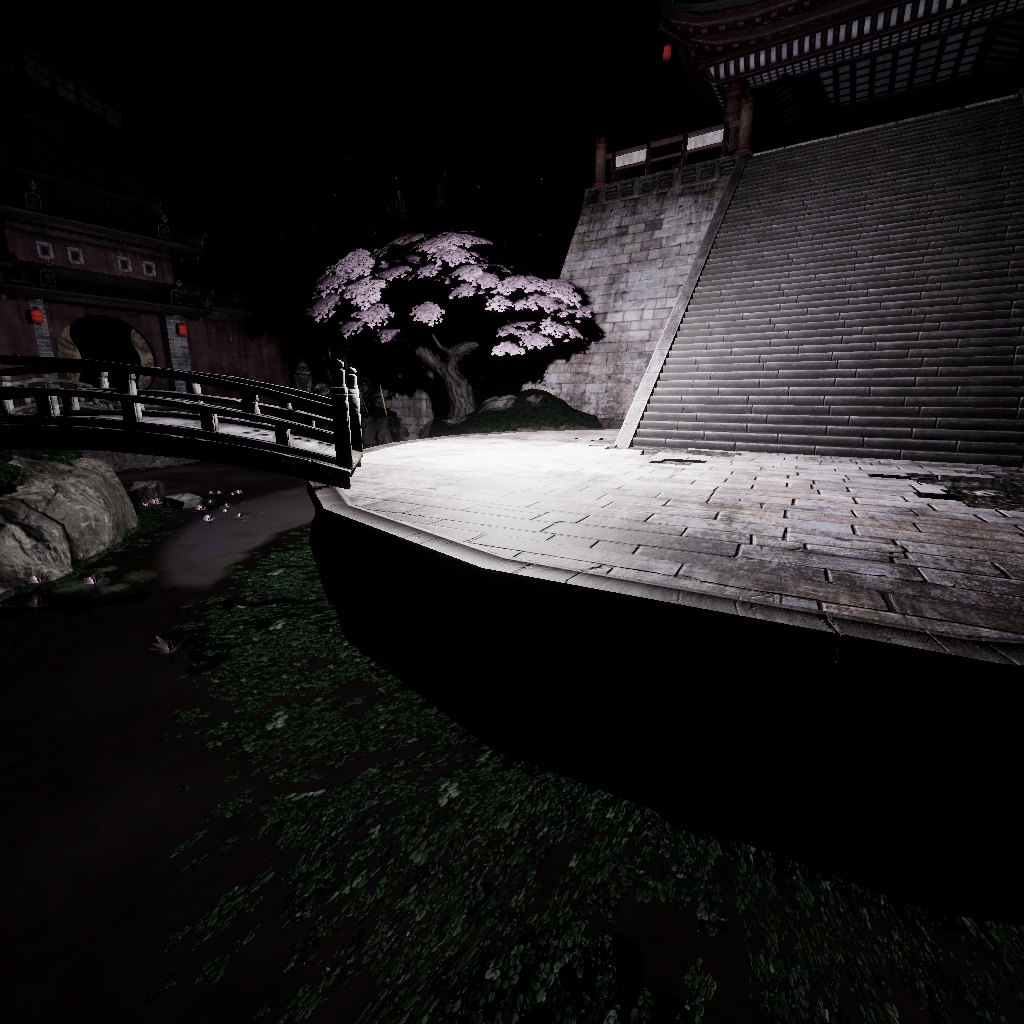}
		\fontsize{8}{12pt}\selectfont PSNR-$15.33$ dB \\SSIM-$0.76$
		\vskip 2pt
		\includegraphics[scale=\scale]{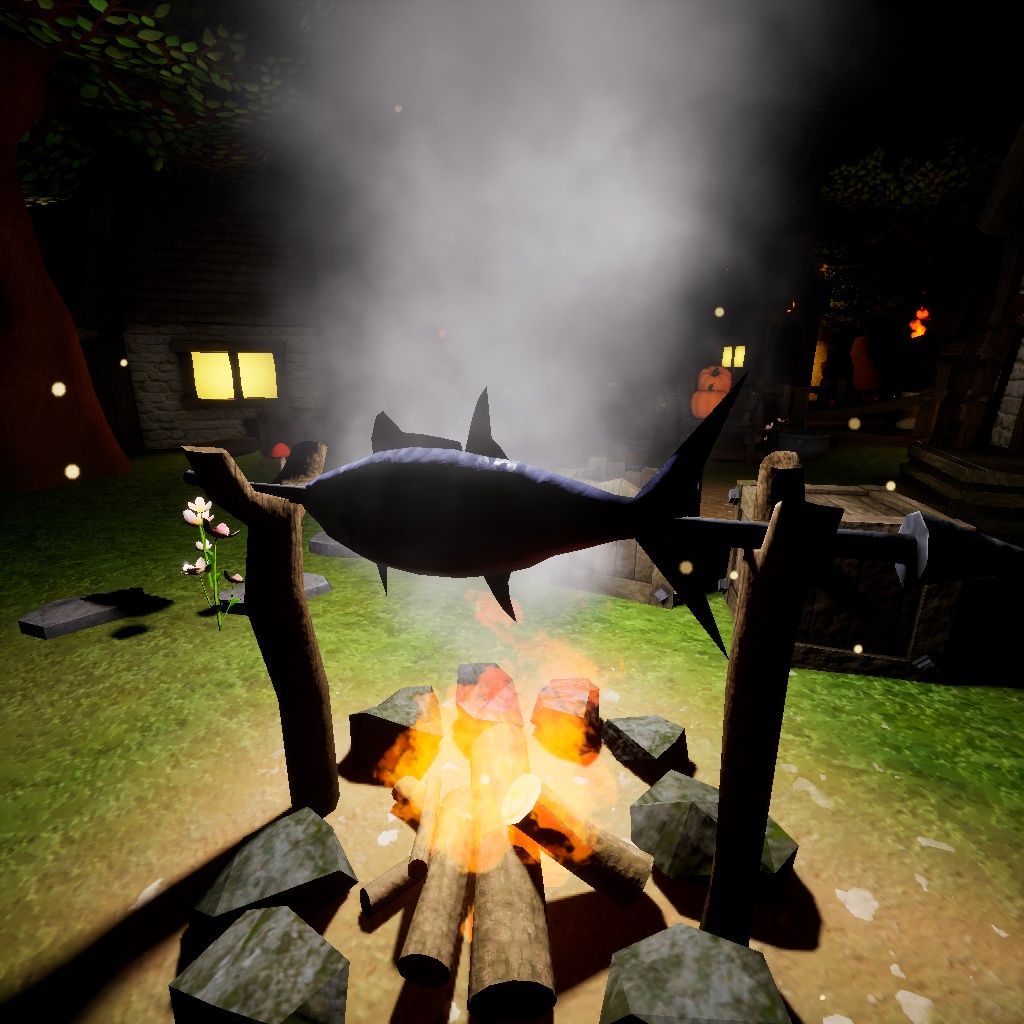}
		\fontsize{8}{12pt}\selectfont PSNR-$13.69$ dB \\SSIM-$0.66$
		\vskip 2pt
		\vskip 2pt
		\includegraphics[scale=\scale]{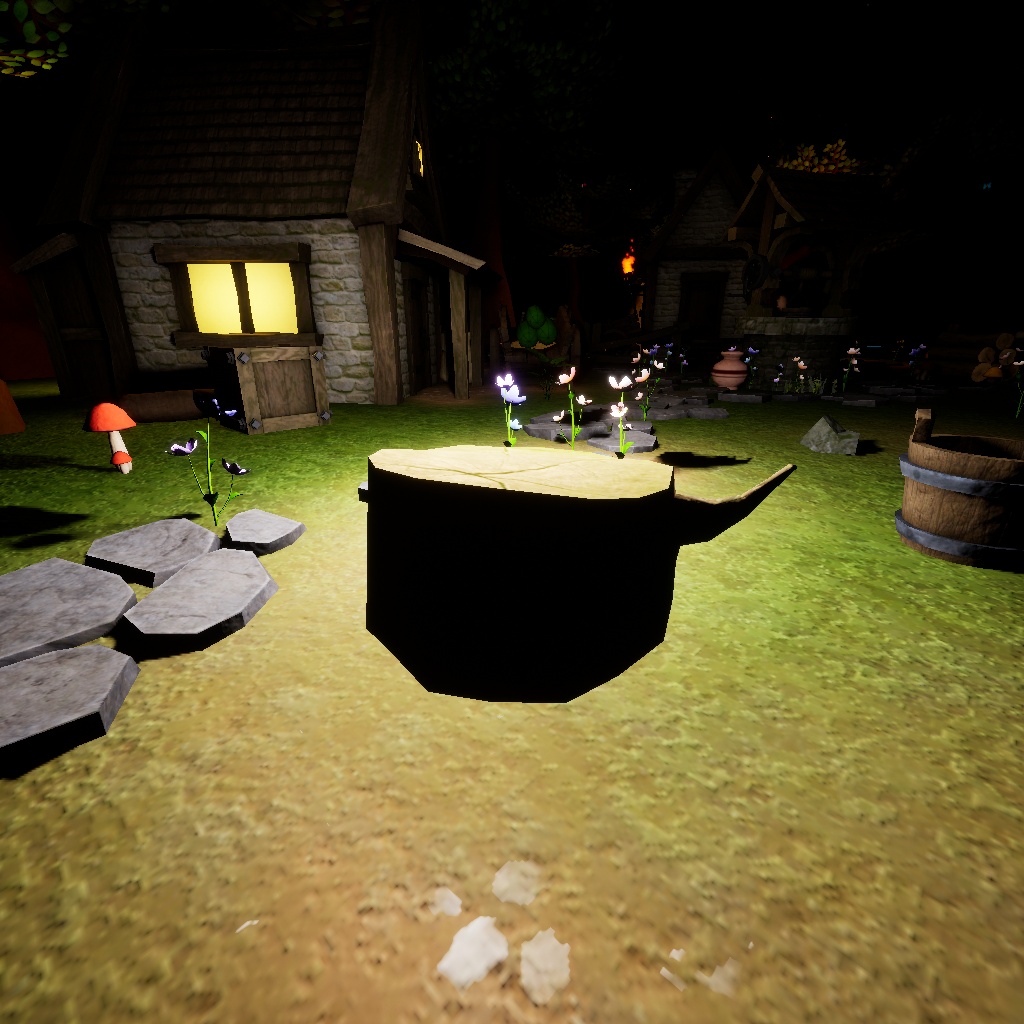}
		\fontsize{8}{12pt}\selectfont PSNR-$14.89$ dB \\SSIM-$0.73$
		\vskip 2pt
        \fontsize{9}{12pt}\selectfont (a) Input \\ {\color{white} (.)}
		\end{center}
  \end{minipage}
  \begin{minipage}{\x\linewidth}
	\begin{center}
	\includegraphics[scale=\scale]{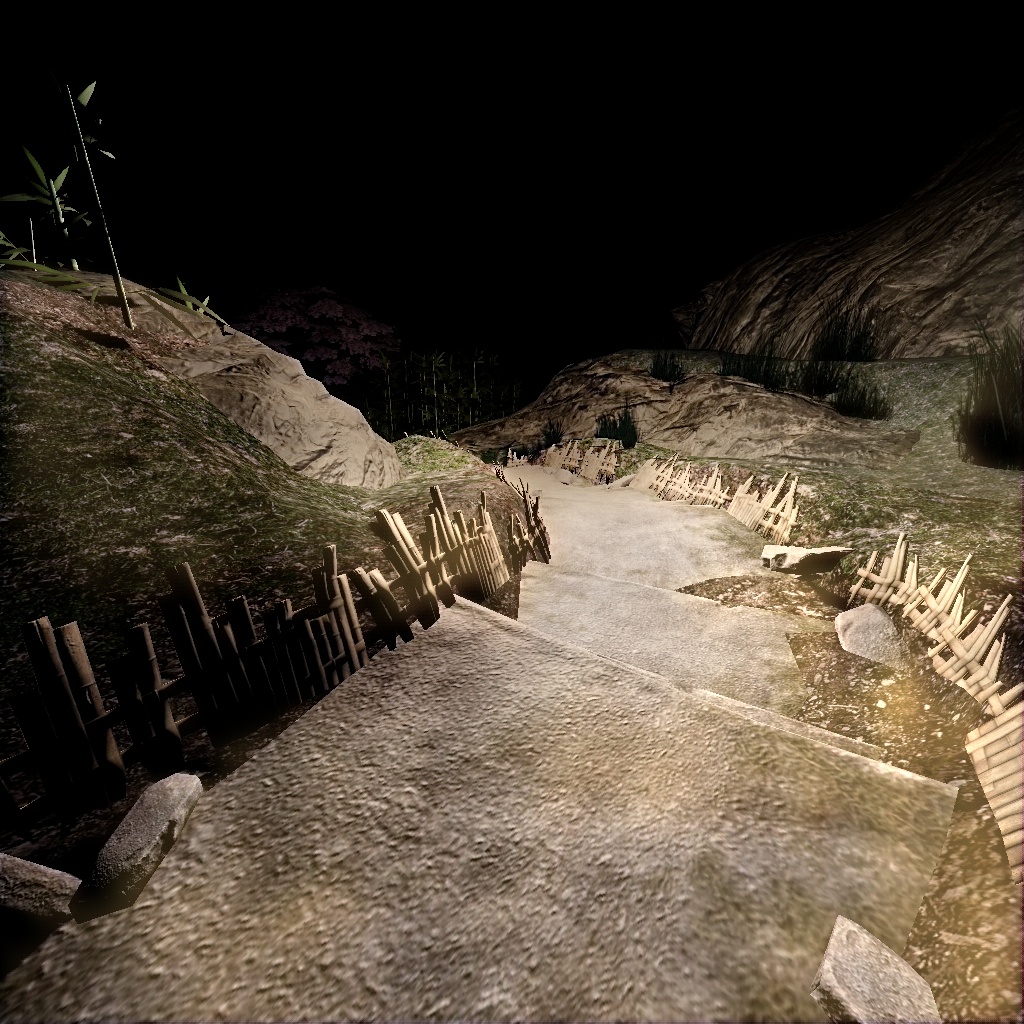}
	\fontsize{8}{12pt}\selectfont PSNR-$14.85$ dB \\SSIM-$0.65$
	\vskip 2pt
	\includegraphics[scale=\scale]{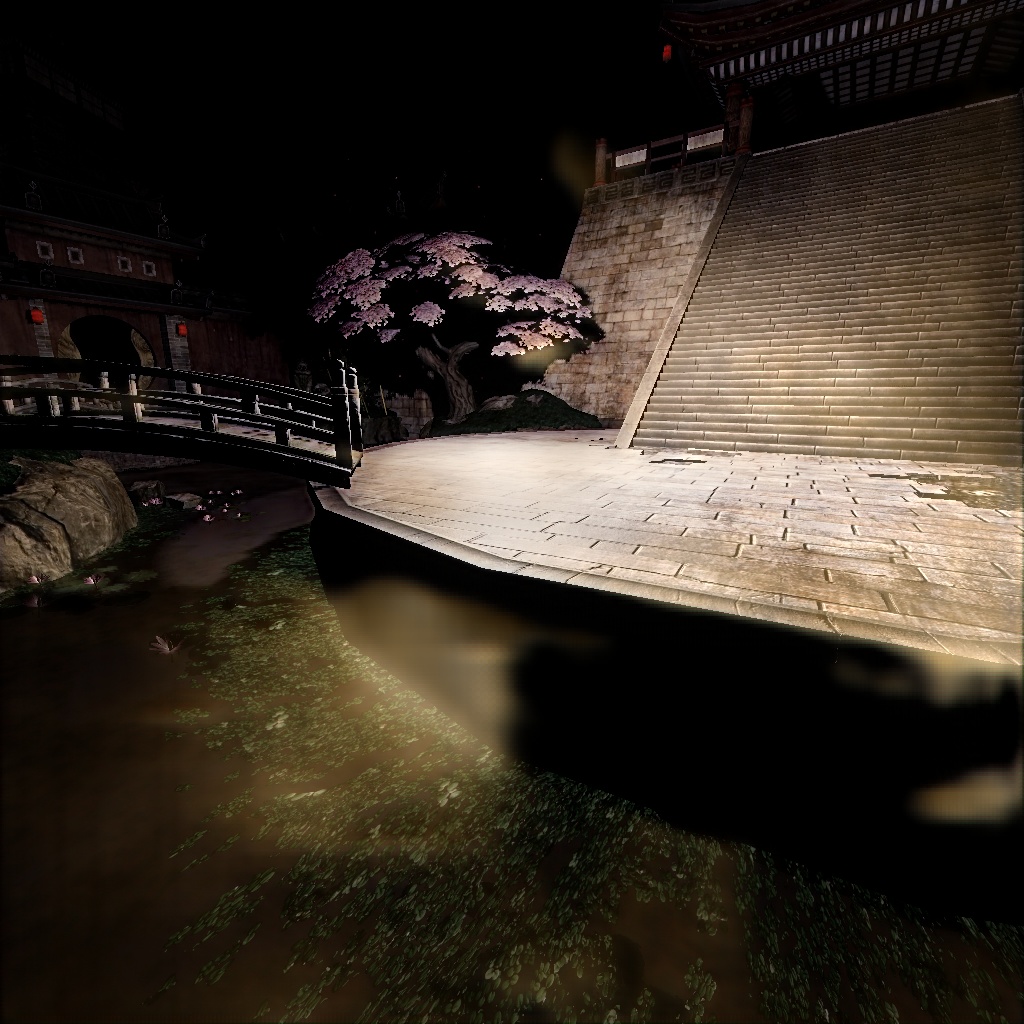}
	\fontsize{8}{12pt}\selectfont PSNR-$14.47$ dB \\SSIM-$0.65$
	\vskip 2pt
	\includegraphics[scale=\scale]{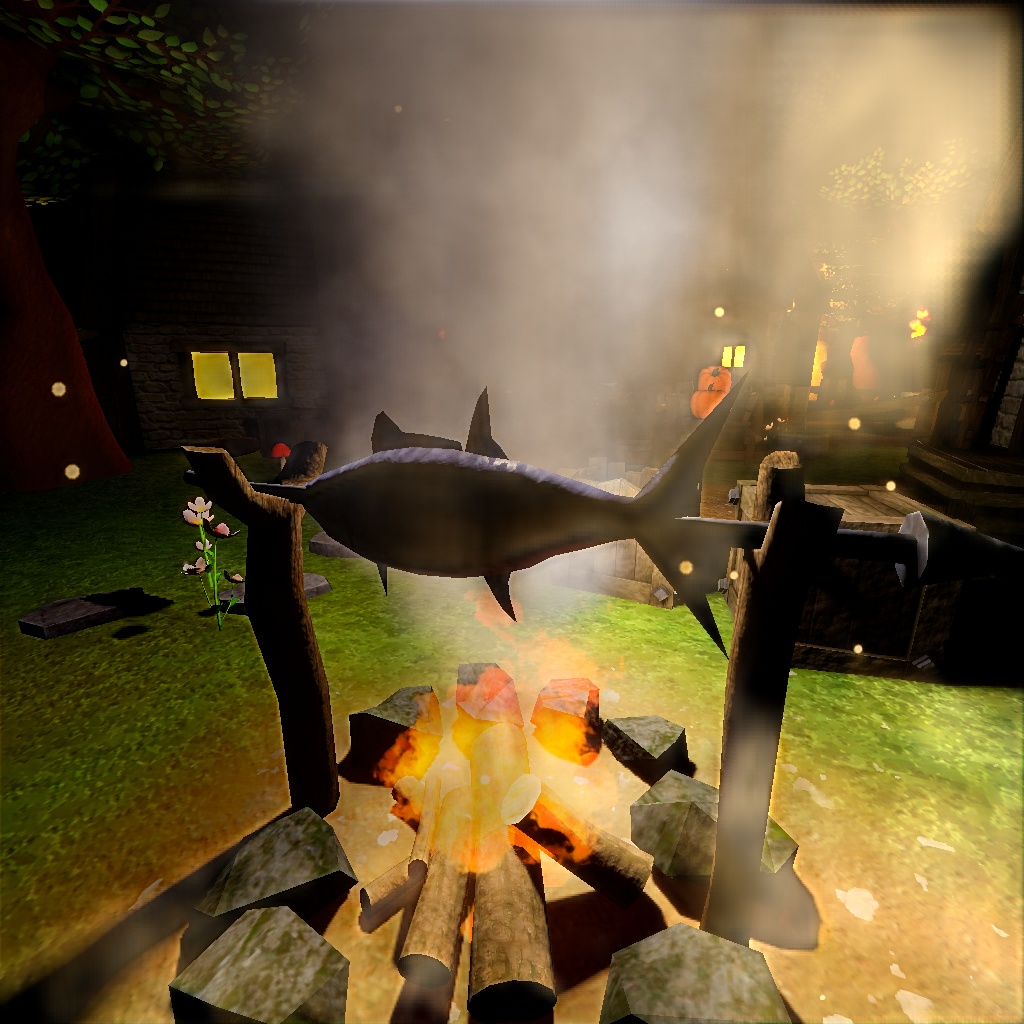}
	\fontsize{8}{12pt}\selectfont PSNR-$14.47$ dB \\SSIM-$0.65$
	\vskip 2pt
    \vskip 2pt
    \includegraphics[scale=\scale]{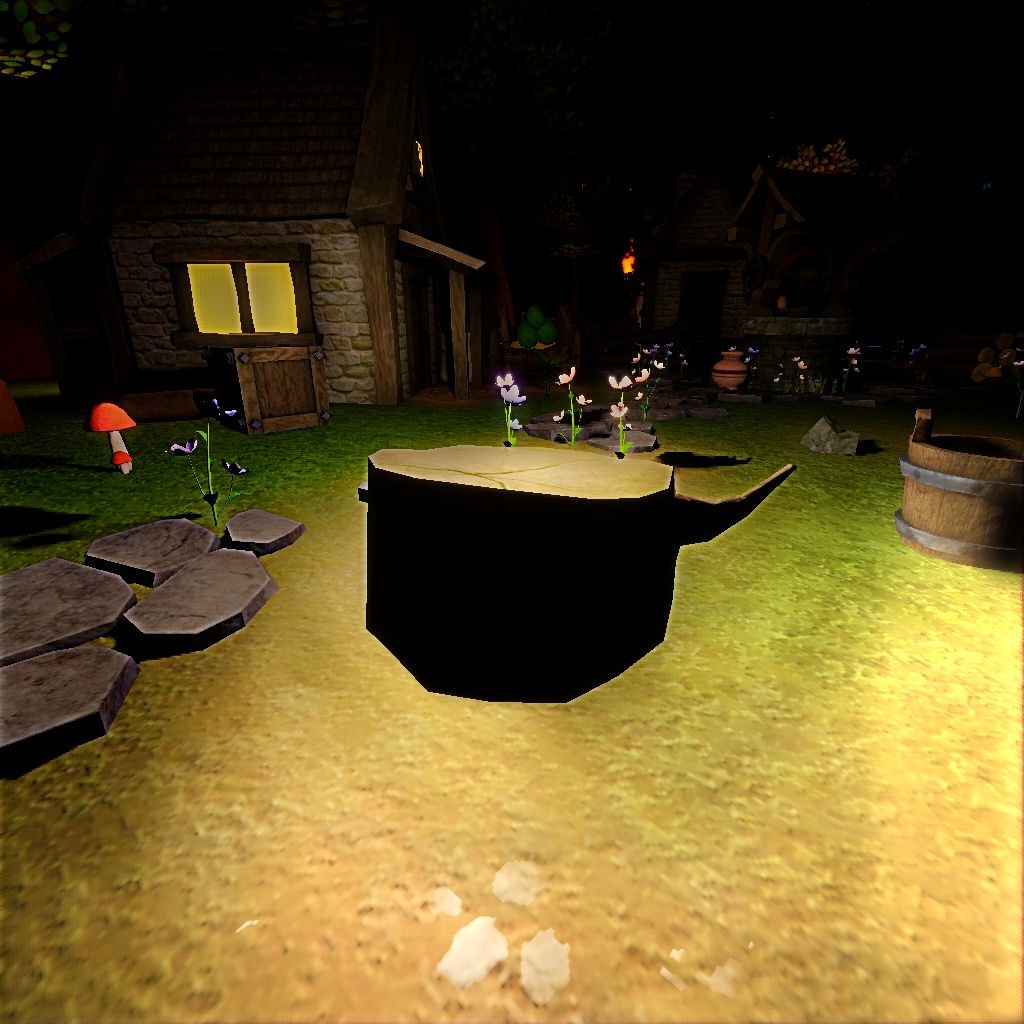}
    \fontsize{8}{12pt}\selectfont PSNR-$14.47$ dB \\SSIM-$0.65$
    \vskip 2pt
	\fontsize{9}{12pt}\selectfont (b) WDRN \\(without gray loss)
  \end{center}
  \end{minipage}
  \begin{minipage}{\x\linewidth}
		\begin{center}
		\includegraphics[scale=\scale]{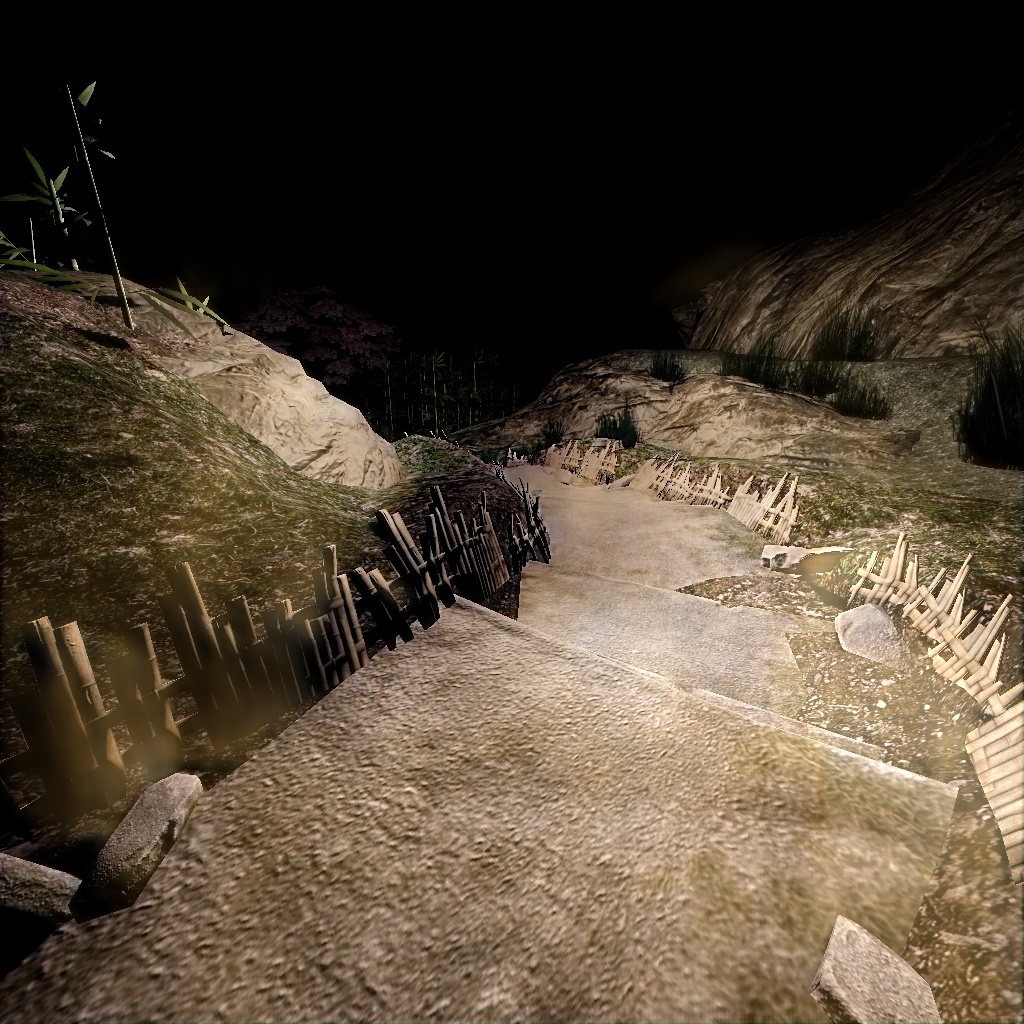}
		\fontsize{8}{12pt}\selectfont PSNR-$14.47$ dB \\SSIM-$0.65$
		\vskip 2pt
		\includegraphics[scale=\scale]{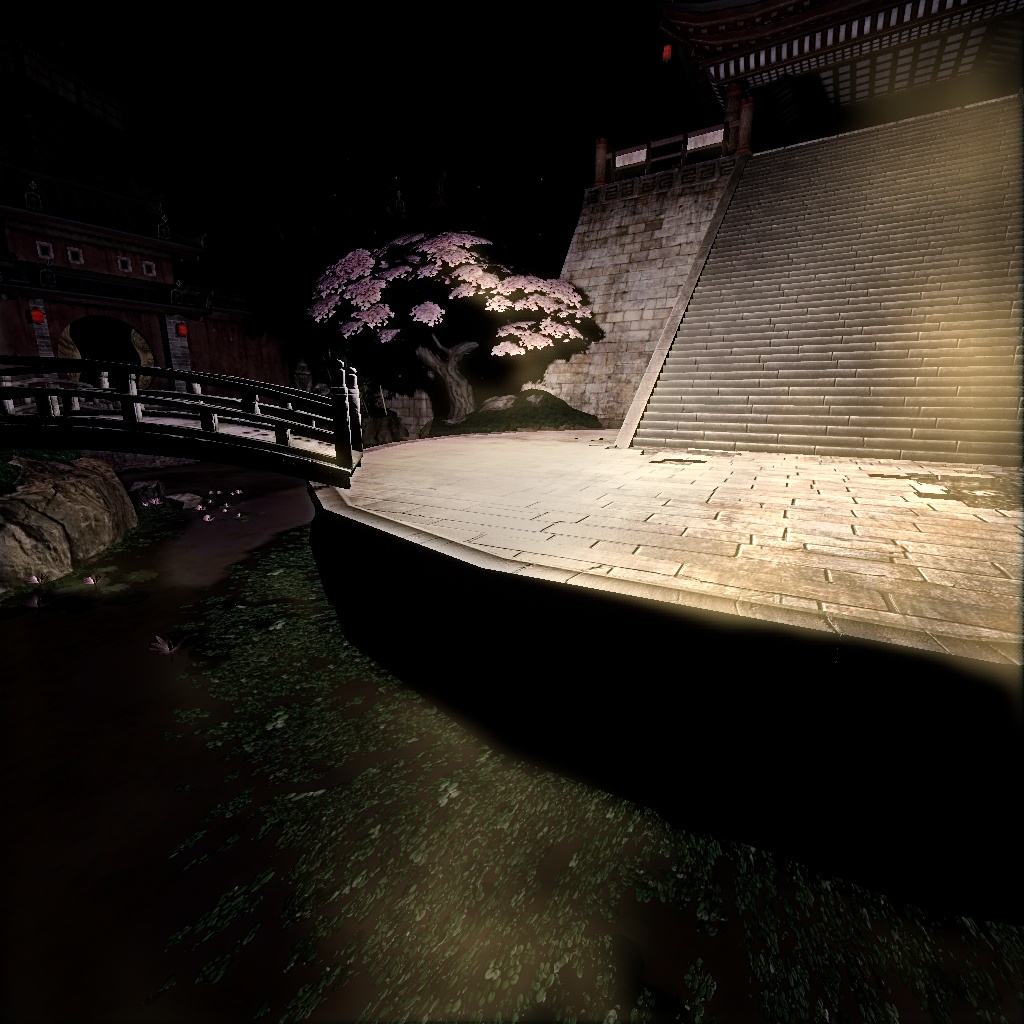}
		\fontsize{8}{12pt}\selectfont PSNR-$16.69$ dB \\SSIM-$0.67$
		\vskip 2pt
		\includegraphics[scale=\scale]{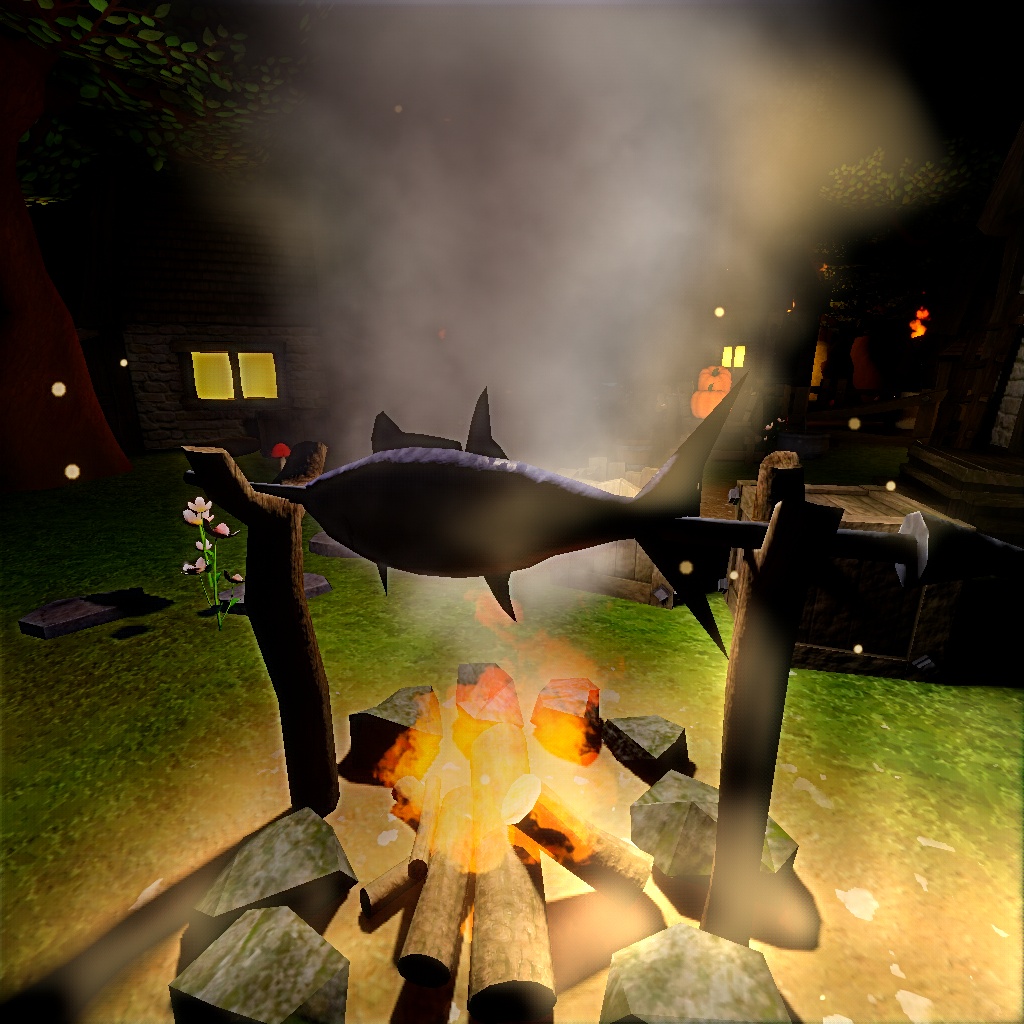}
		\fontsize{8}{12pt}\selectfont PSNR-$12.46$ dB \\SSIM-$0.63$
		\vskip 2pt
        \vskip 2pt
        \includegraphics[scale=\scale]{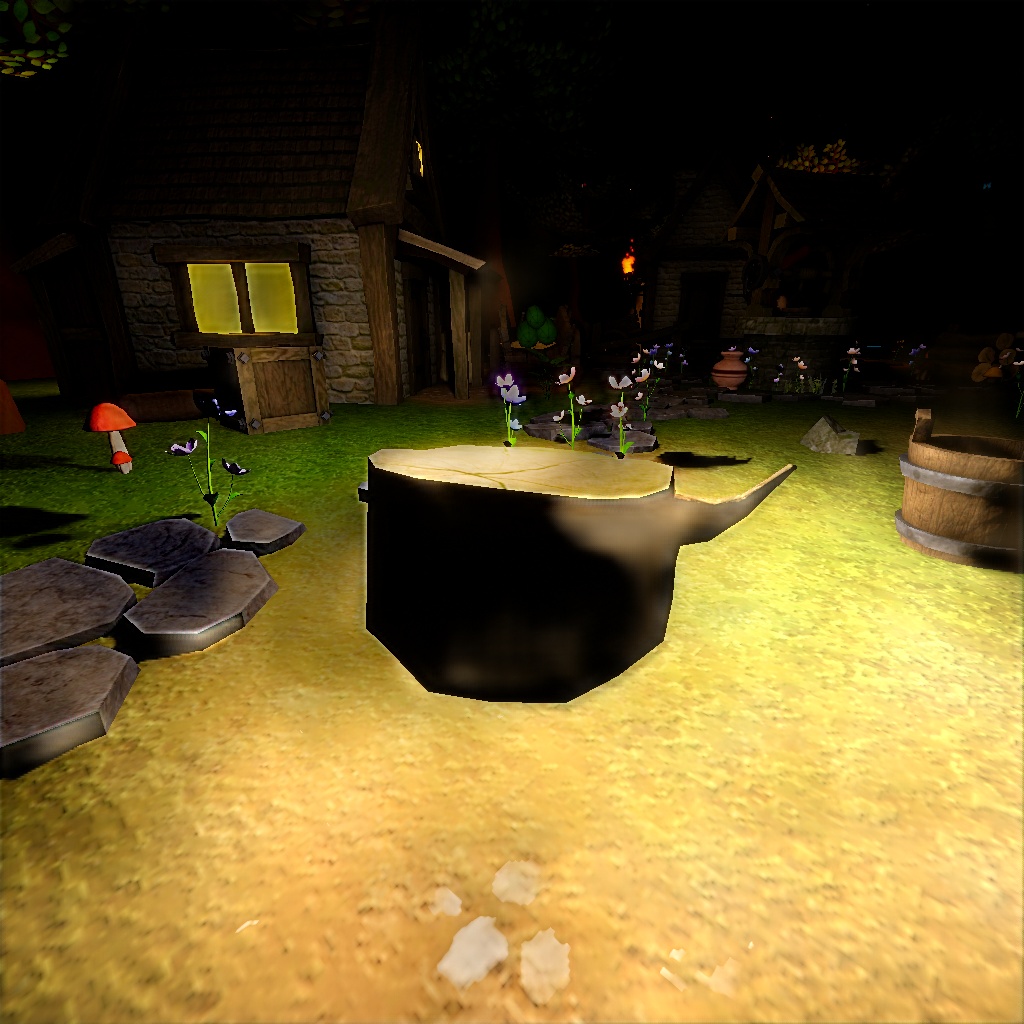}
        \fontsize{8}{12pt}\selectfont PSNR-$19$ dB \\SSIM-$0.79$
        \vskip 2pt
		\fontsize{9}{12pt}\selectfont (c) WDRN \\(with gray loss)
		\end{center}
  \end{minipage}
  \begin{minipage}{\x\linewidth}
		\begin{center}
		\includegraphics[scale=\scale]{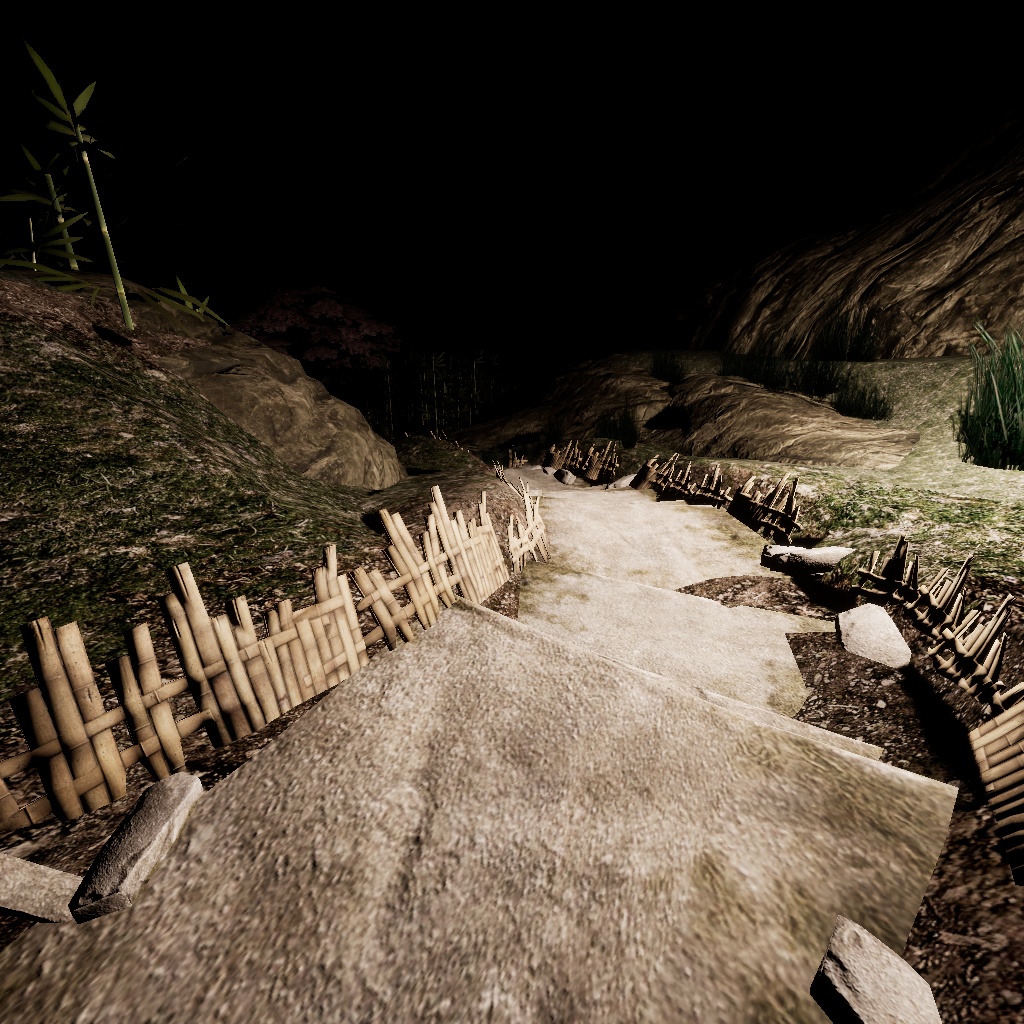}
		\fontsize{8}{12pt}\selectfont {\color{white} .\\.}
		\vskip 2pt
		\includegraphics[scale=\scale]{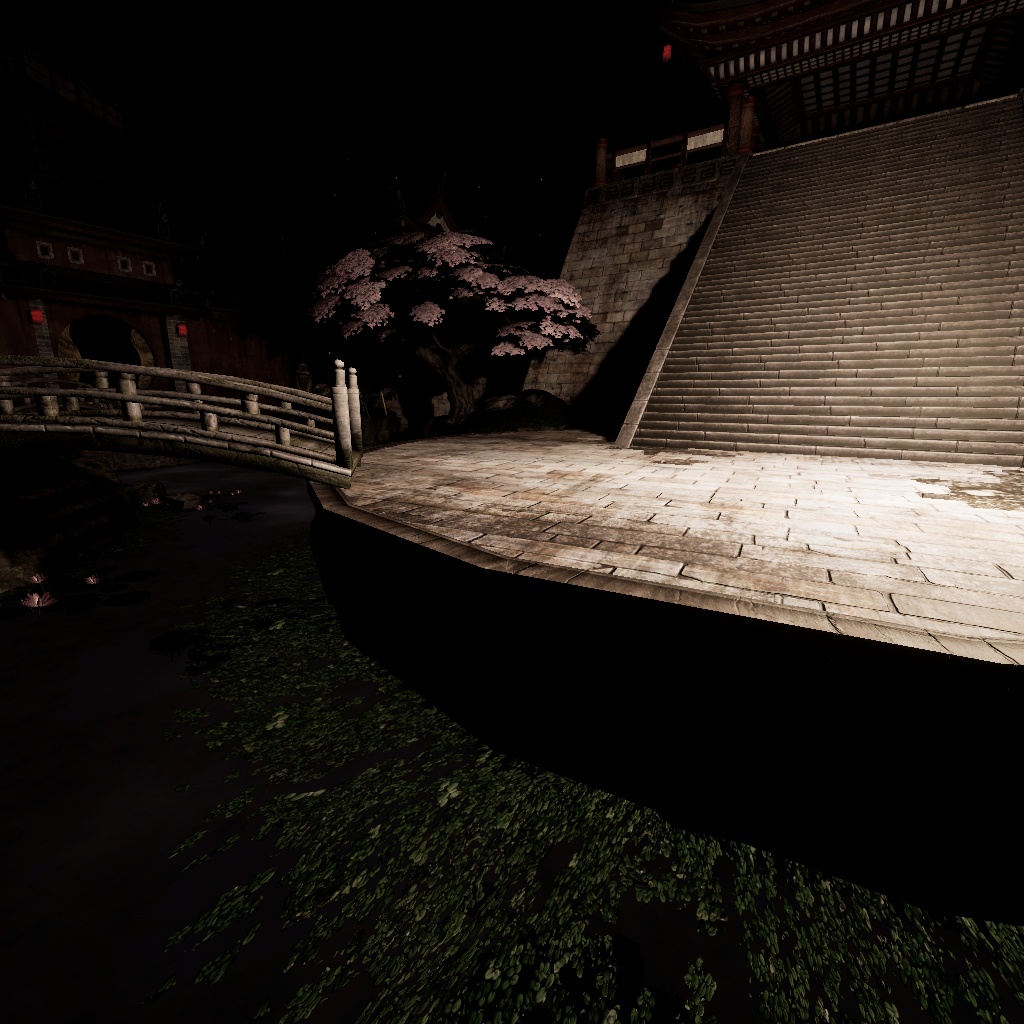}
		\fontsize{8}{12pt}\selectfont {\color{white} .\\.}
		\vskip 2pt
		\includegraphics[scale=\scale]{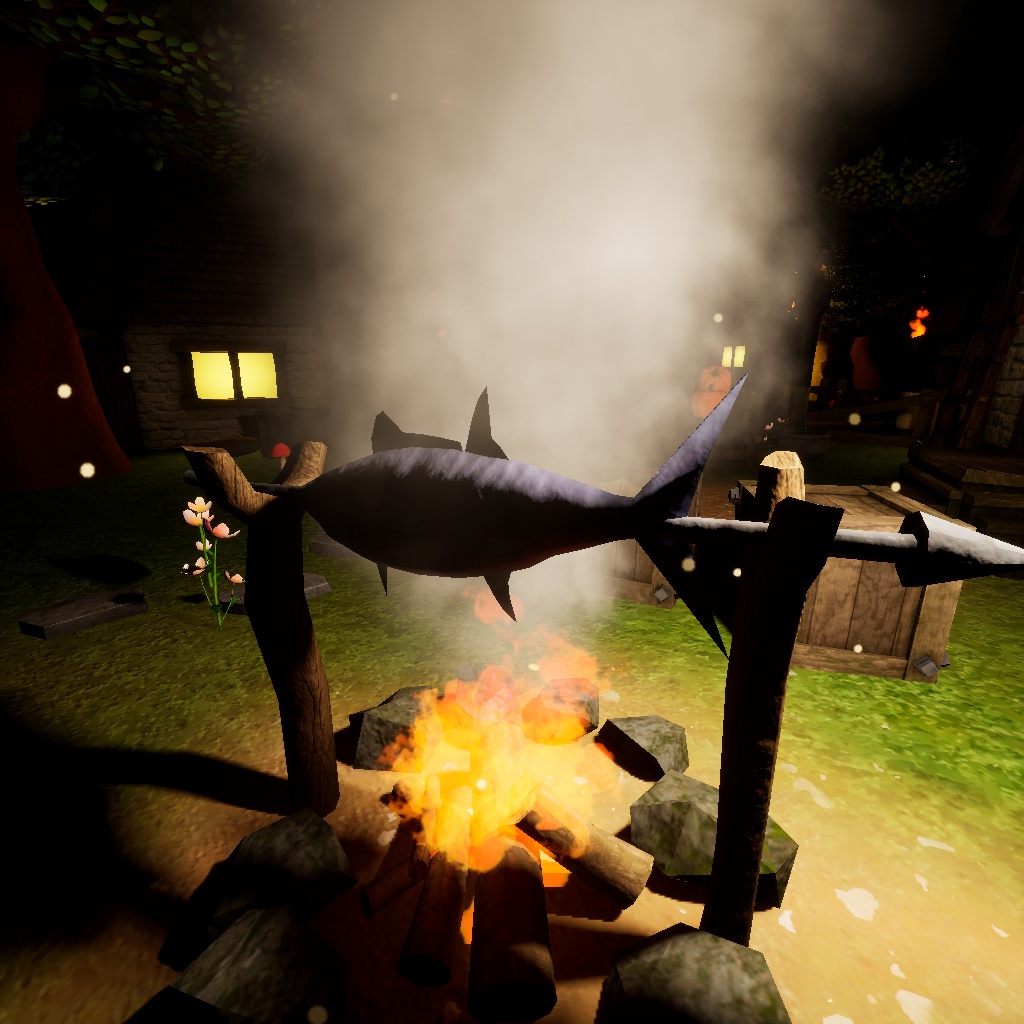}
		\fontsize{8}{12pt}\selectfont {\color{white} .\\.}
		\vskip 2pt
        \vskip 2pt
        \includegraphics[scale=\scale]{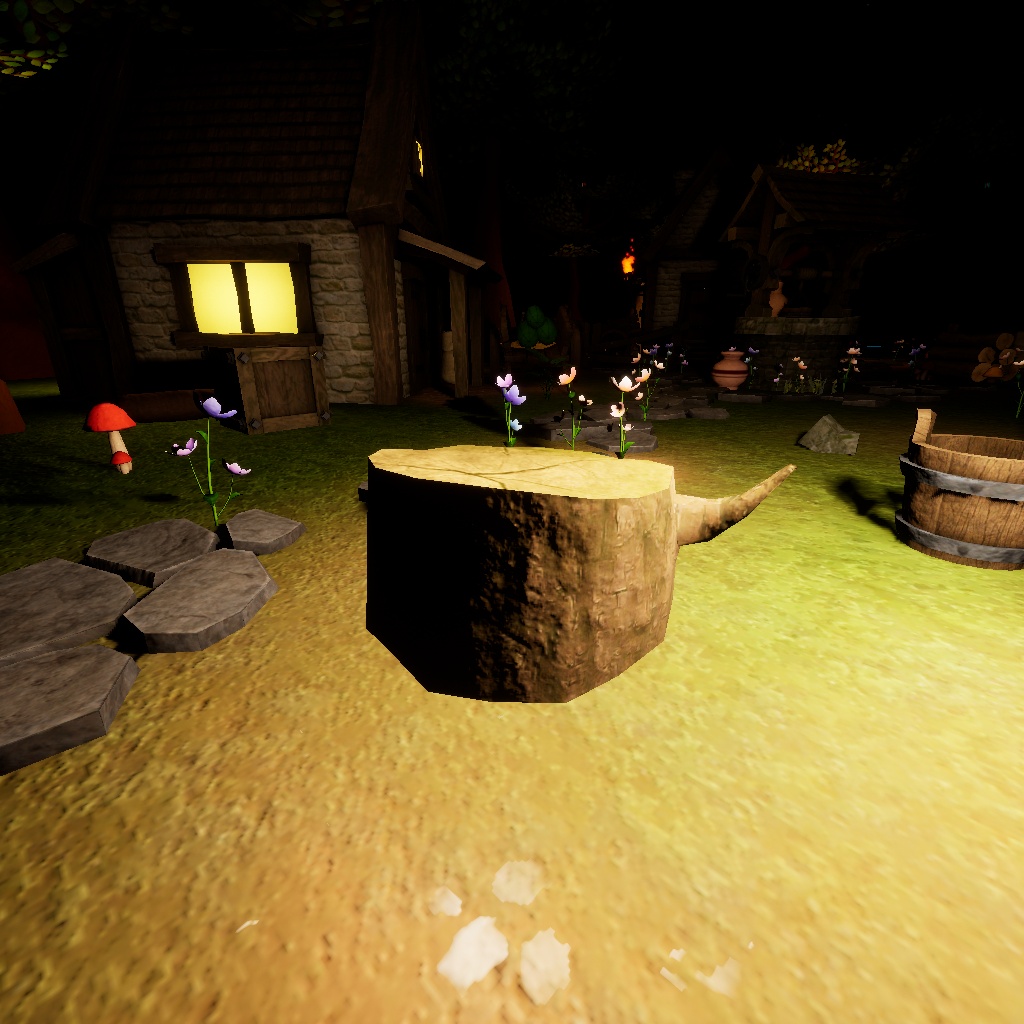}
        \fontsize{8}{12pt}\selectfont {\color{white} .\\.}
        \vskip 2pt
		\fontsize{9}{12pt}\selectfont (d) Ground truth \\ {\color{white} (.)}
		\end{center}
  \end{minipage}
\caption{Sample results with the proposed method on the validation set of VIDIT. From the visual inspection of Fig. a, b, c and d, it is evident that the proposed WDRN architecture is able to capture the colour temperature of the target domain to a large extent. Although WDRN trained with and without gray loss have comparable performance in terms of quantitative metrics, the former has superior visual quality.}
\label{fig:sample_results_one-one-relight}
\end{figure}

Fig. \ref{fig:sample_results_one-one-relight} shows relit examples corresponding to four input images from VIDIT validation set for one-to-one relighting problem using the proposed WDRN trained with and without gray loss. For certain cases, the input images have better PSNR and/or SSIM than the relit images although the latter is perceptually closer to the ground truth. As evident from the figure, the position of the illuminant $\theta_2$ and the color temperature $T_2$ of the target image is predicted with considerable visual similarity with the WDRN architecture. However, WDRN failed to inpaint information in the shadows that should have been uncovered with the change in illuminant position. Similarly, WDRN failed to recast the shadows in the target domain image. These two issues can be assumed to be the biggest challenges in relighting problem. Although WDRN trained with and without gray loss have comparable performance in terms of quantitative metrics, they differ to a certain extent visually. While WDRN with gray loss inpainted some information in shadowy areas in rows 1 and 4 of the figure, WDRN without gray loss failed to do so. Similarly, WDRN without gray loss inpainted information in unwanted regions in rows 2 and 4 of the figure, thereby having an incorrect representation of the target domain image.

\begin{table}[h!]
\centering
\caption{Performance comparison of WDRN with competing entries in scene relighting and illumination estimation challenge, track-1 one-to-one relighting at AIM 2020 workshop. The MPS, used to determine the final ranking, is computed following Eq.~\eqref{eq:MPS}.}
\resizebox{\textwidth}{!}{%
\begin{tabular}{l||c|c|c|c|c|}
Team &\textbf{MPS}&SSIM&LPIPS&PSNR&Run-time\\
\hline\hline
\textbf{Our method(with gray loss)}&\textbf{0.6452} \textbf{(1)}&\textbf{0.6310} (2)&\textbf{0.3405} \textbf{(1)}&\textbf{17.0717} (2)&\textbf{{}0.03s}\\
\textbf{Our method(without gray loss)}&\textbf{0.6451} (2)&\textbf{0.6362} \textbf{(1)}&\textbf{0.3460} (3)&\textbf{16.8927} (6)&\textbf{0.03s}\\
Team 2&0.6436 (3)&0.6301 (3)&0.3430 (2)&16.6801 (8)&13s\\
Team 3&0.6216 (4)&0.6091 (4)&0.3659 (5)&16.8196 (7)&6s\\
Team 4&0.5897 (5)&0.5298 (7)&0.3505 (4)&17.0594 (3)&0.04s\\
Team 5&0.5892 (6)&0.5928 (6)&0.4144 (7)&17.4252 \textbf{(1)}&0.5s\\
Team 6&0.5603 (7)&0.5236 (8)&0.4029 (6)&16.5136 (9)&0.01s\\
Team 7&0.5339 (8)&0.5666 (6)&0.4989 (8)&16.9234 (4)&0.006s\\
Team 8&0.3746 (9)&0.3769 (9)&0.6278 (9)&16.8949 (5)&0.12s\\
\end{tabular}}
\label{tab:table1}
\end{table}

\par
Table \ref{tab:table1} shows the performance comparison of the proposed WDRN with other competing entries in one-to-one relighting challenge of AIM 2020 workshop. We proposed two variants of WDRN in the competition - one trained with gray loss and the other without it. Both variants were able to achieve better MPS than other methodologies. While the method with gray loss obtained the highest MPS of $0.6452$, the one without it obtained highest score in SSIM and $2^{nd}$ second highest MPS score of $0.6451$. Organizers merged both the methods as the architecture followed was same even though the loss functions used were different. Additionally, owing to the network lightness, runtime of WDRN is considerably lower than the immediate runner-ups. 

\par
As future work, the proposed WDRN network can be modified to address the related problems like any-to-any relighting, under-exposure correction etc. In any-to-any relighting, WDRN may be modified to feature an additional encoder to which a guide image from the target domain can be given as an additional input and the illumination properties of the guide image can be injected into the input image in the encoder section itself. Another way to realise this is to modify WDRN to to have additional scalar inputs corresponding to the target illumination settings and integrate them into encoder in a manner demonstrated in \cite{relightnet_a}.

\section{Ablation Studies}
\label{section:ablation_studies}

\subsection{Wavelet Domain Network}
\label{subsection:wavelet_ablation}
To find out the effectiveness of wavelet decomposition approach, an ablation study was conducted using a 3-level encoder-decoder network. In the equivalent pixel domain network, wavelet decomposition was replaced with convolutional downsampling with a stride of two. Similarly, wavelet interpolation has been replaced with transposed convolution with an upscale factor of two. The results of the experiment on validation set of VIDIT is reported in Table \ref{tab:table_ablation_pixel}. It is conclusive that wavelet domain network obtained superior performance in terms of all evaluation metrics of the experiment which proves its effectiveness.

\begin{table}[h!]
\centering
\caption{Ablation study of wavelet domain network}
\begin{tabular}{l||c|c|c|c|}
Domain &\textbf{MPS}&SSIM&LPIPS&PSNR \\
\hline\hline
Pixel & 0.6918 & 0.6619 & 0.2783 & 17.3934 \\
\textbf{Wavelet} & \textbf{0.6935} & \textbf{0.6642} & \textbf{0.2771} & \textbf{17.4539}\\
\end{tabular}
\label{tab:table_ablation_pixel}
\end{table}

\subsection{Wavelet Decomposition Levels}
\label{subsection:decomposition_levels_ablation}
To investigate the effect of various levels of wavelet decomposition, an extensive ablation study have been carried out. Experiments were conducted with two, three and four levels of wavelet decomposition in the encoder-decoder architecture. The model training was limited to 60 iterations to avoid the risk of overfitting. Table \ref{tab:table_ablation_levels} shows the comparison of the performance obtained with different decomposition levels on the validation set of VIDIT dataset. Notably, the 3-level decomposed network shows superior performance in terms of perceptual metrics like SSIM and LPIPS while the 4-level network achieves highest PSNR.

\begin{table}[h!]
\centering
\caption{Ablation study of wavelet decomposition levels}
\begin{tabular}{l||c|c|c|c|}
Decomposition level &\textbf{MPS}&SSIM&LPIPS&PSNR \\
\hline\hline
2 level & 0.6842 & 0.6486 & 0.28 & 17.0962\\
\textbf{3 level} & \textbf{0.6935} & \textbf{0.6642} & \textbf{0.2771} & 17.4539\\
4 level & 0.6908 & 0.661  & 0.2792 & \textbf{17.5586}\\
\end{tabular}
\label{tab:table_ablation_levels}
\end{table}

\section{Conclusions}
\label{section:conclusions}
In this paper we proposed a novel multi-resolution encoder-decoder network employing wavelet based decomposition called wavelet decomposed RelightNet to address one-to-one image relighting problem. Additionally, a novel gray loss term tailored for the problem resulted in visually superior relit images. The experimental results have proved the effectiveness of the proposed WDRN both qualitatively and in terms of various quantitative parameters. The proposed WDRN can be modified to address other related problems like any-to-any relighting, under-exposure correction etc.

\section*{Acknowledgements}
 We gratefully acknowledge the support of NVIDIA PSG Cluster and Trivandrum Engineering Science and Technology Research Park (TrEST) in providing the computational resource to conduct this research.

\clearpage
%
%
\bibliographystyle{splncs04}
\bibliography{main}
\end{document}